\documentclass[letterpaper]{article} 
\usepackage{aaai25}  
\usepackage{times}  
\usepackage{helvet}  
\usepackage{courier}  
\usepackage[hyphens]{url}  
\usepackage{graphicx} 
\usepackage{natbib}  
\usepackage{caption} 
\usepackage{algorithm}
\usepackage{algorithmic}

\usepackage{bm}
\usepackage{multirow}
\usepackage{bbding}
\usepackage{pifont}
\usepackage{url}
\usepackage{amsmath}
\usepackage{xcolor}         
\usepackage{multirow}
\usepackage{booktabs}
\usepackage{amssymb}

\newcommand{\methodNameSpace}{GraphAvatar }
\usepackage{newfloat}
\usepackage{listings}
\DeclareCaptionStyle{ruled}{labelfont=normalfont,labelsep=colon,strut=off} 
\floatstyle{ruled}
\newfloat{listing}{tb}{lst}{}
\floatname{listing}{Listing}
\title{GraphAvatar: Compact Head Avatars with GNN-Generated 3D Gaussians}
\author{
    Xiaobao Wei\textsuperscript{\rm 1,2},
    Peng Chen\textsuperscript{\rm 1,2}, 
    Ming Lu\textsuperscript{\rm 3},
    Hui Chen\textsuperscript{\rm 1}\thanks{Corresponding Author.},
    Feng Tian\textsuperscript{\rm 1}
}
\affiliations{
    \textsuperscript{\rm 1}Institute of Software, Chinese Academy of Sciences\\
    \textsuperscript{\rm 2}University of Chinese Academy of Sciences \\
    \textsuperscript{\rm 3}Intel Labs China\\

    weixiaobao0210@gmail.com
}

\usepackage{bibentry}

\begin{document}

\maketitle


\begin{abstract}
Rendering photorealistic head avatars from arbitrary viewpoints is crucial for various applications like virtual reality. Although previous methods based on Neural Radiance Fields (NeRF) can achieve impressive results, they lack fidelity and efficiency. Recent methods using 3D Gaussian Splatting (3DGS) have improved rendering quality and real-time performance but still require significant storage overhead. In this paper, we introduce a method called GraphAvatar that utilizes Graph Neural Networks (GNN) to generate 3D Gaussians for the head avatar. Specifically, GraphAvatar trains a geometric GNN and an appearance GNN to generate the attributes of the 3D Gaussians from the tracked mesh. Therefore, our method can store the GNN models instead of the 3D Gaussians, significantly reducing the storage overhead to just 10MB. To reduce the impact of face-tracking errors, we also present a novel graph-guided optimization module to refine face-tracking parameters during training. Finally, we introduce a 3D-aware enhancer for post-processing to enhance the rendering quality. We conduct comprehensive experiments to demonstrate the advantages of GraphAvatar, surpassing existing methods in visual fidelity and storage consumption. The ablation study sheds light on the trade-offs between rendering quality and model size. The code will be released at: https://github.com/ucwxb/GraphAvatar
\end{abstract}

%
\begin{links}
    \link{Code}{https://github.com/ucwxb/GraphAvatar}
\end{links}

\section{Introduction}
Rendering photorealistic head avatars from any viewpoint is essential for virtual reality and augmented reality applications. Key aspects such as visual fidelity, rendering speed, and storage overhead are crucial. With the advancement of deep learning, methods based on neural fields have become prevailing due to their advantages in these aspects. The seminal neural field work Neural Radiance Fields (NeRF)~\cite{mildenhall2021nerf} and its variants~\cite{wang2021neus, wei2023noc} have achieved impressive results for neural rendering and reconstruction. 

Regarding head avatars,   
NeRF-based methods~\cite{Gao2022nerfblendshape, Zielonka2022InstantVH, zheng2023pointavatar, lombardi2019neural, lombardi2021mixture} have focused on improving the generation from short RGB video inputs. Based on 3D morphable models (3DMM)~\cite{li2017learning,paysan20093d}, these methods learn neural fields to create 3D-consistent and interpretable digital head avatars. These approaches enable high-quality rendering and diverse applications, such as facial retargeting, expression editing, and rapid avatar generation. Although point~\cite{zheng2023pointavatar} and hash table ~\cite{Zielonka2022InstantVH, Gao2022nerfblendshape} representations have been explored for acceleration, they still struggle with the excessive sampling in the volume rendering of NeRF, which prevent them from achieving real-time rendering easily. Moreover, due to the implicit characteristics of NeRF, these methods lack controllability and do not generalize well to the novel poses and expressions.

Recently, 3D Gaussian Splatting~\cite{kerbl20233d} successfully represents a static scene as 3D Gaussians and renders with a differentiable rasterizer, significantly accelerating novel view synthesis. Due to the unstructured nature of 3D Gaussian representation, 3DGS also excels in controllability and generalization compared to NeRF~\cite{fang2023gaussianeditor}. Many recent methods~\cite{qian2024gaussianavatars, xiang2024flashavatar, xu2023gaussianheadavatar, dhamo2023headgas, wei2024gazegaussian, chen2024mixed} have been proposed to apply 3DGS to head avatars. Based on face tracking parameters, these methods initialize or bind 3D Gaussians to the geometry prior using neutral mesh~\cite{xu2023gaussianheadavatar}, tracked mesh~\cite{qian2024gaussianavatars}, or UV mapping~\cite{xiang2024flashavatar}. Subsequently, the 3D Gaussians are learned from short RGB video inputs involving densification, such as cloning and pruning, and deformations from canonical space. These methods leverage 3D head meshes to incorporate prior knowledge, facilitating better convergence of 3D Gaussians, which capture human heads' appearance, geometry, and dynamics.

\begin{figure*}[t]
    \centering
    \includegraphics[width=1.0\textwidth]{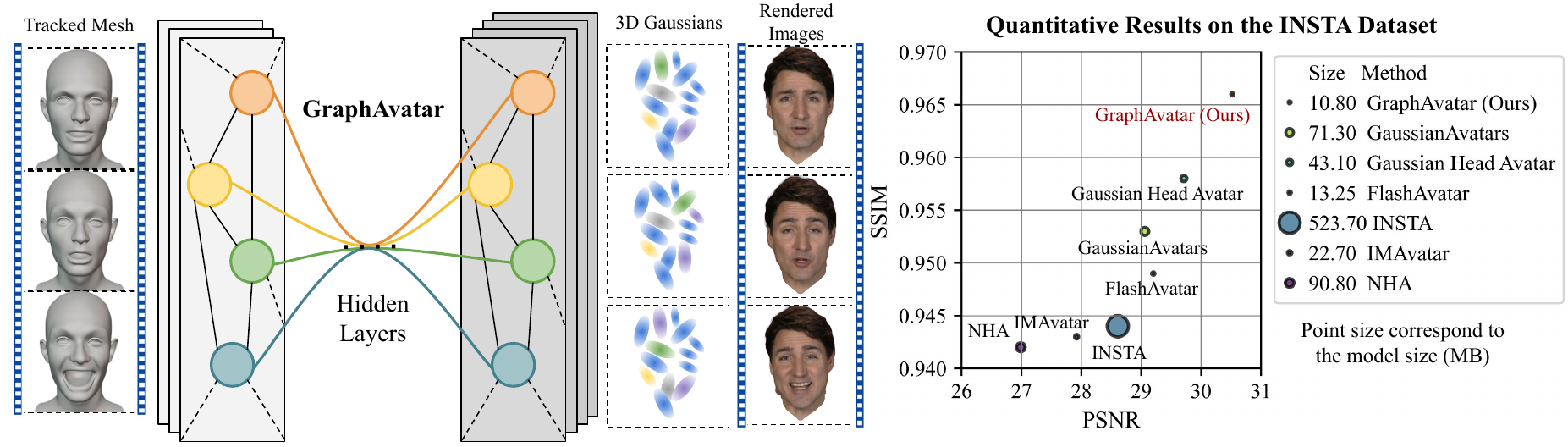}
    \vspace{-2mm}
    \caption{GraphAvatar leverages graph neural networks to generate 3D Gaussians, which are then rasterized into high-fidelity images based on tracked meshes. Compared to contemporary approaches, GraphAvatar not only delivers superior rendering performance but also features the most compact model size, substantially minimizing storage requirements.}
    \label{fig:teaser}
    \vspace{-5mm}
\end{figure*}

However, existing 3DGS-based approaches still face two obvious limitations: 
1) \textbf{Significant and fluctuating storage overhead.} Since the 3D head mesh contains many triangles, assigning multiple 3D Gaussians to each triangle results in excessive storage overhead due to the large number of triangles in the 3D head mesh. Furthermore, the number of 3D Gaussians also changes during the densification process. This leads to a fluctuating storage overhead, posing challenges for practical applications.
2) \textbf{Heavy reliance on face tracking.} Since these methods initialize or bind 3D Gaussians to the tracked head mesh, they heavily rely on the accuracy of face tracking. The accumulated errors of face tracking will impact the training of 3D Gaussians.   

To address these issues, we propose a novel method named GraphAvatar, which utilizes Graph Neural Networks (GNN)~\cite{kipf2016semi, ranjan2018generating} to generate 3D Gaussians for photorealistic head avatars. As shown in Fig.~\ref{fig:teaser}, GraphAvatar optimizes a geometric GNN and an appearance GNN to generate 3D Gaussians using tracked meshes as input. These 3D Gaussians serve as anchors and are fed into a view-dependent MLP to learn 3D Gaussian offsets related to different viewpoints. The predicted offsets adjust the anchor 3D Gaussians, breaking free from the constraints imposed by the tracked meshes, allowing for learning better details. Subsequently, rasterization is used to render the adjusted 3D Gaussians into photorealistic head avatars. Thus, our method significantly reduces the storage overhead by storing the GNN models instead of many 3D Gaussians, making our method more compact. To reduce the impact of face-tracking errors, we also introduce an advanced graph-guided optimization module to optimize face-tracking parameters during the training. Finally, to reduce the over-smoothing induced by GNN, we incorporate a lightweight 3D-aware enhancer for post-processing, which utilizes rendered depth maps to improve rendering quality. Our main contributions are summarized as follows:

\begin{itemize}
\item We propose GraphAvatar, a novel and compact method that utilizes GNN to generate 3D Gaussians. Our method only stores the GNN models, eliminating the need to store 3D Gaussians directly.

\item To reduce reliance on tracked meshes, we introduce a graph-guided optimization module to refine face-tracking parameters during training.

\item To alleviate the over-smoothing induced by GNN, we incorporate a 3D-aware enhancer to enhance details for the rendered images.

\item We conduct comprehensive experiments on challenging datasets to demonstrate that GraphAvatar achieves state-of-the-art rendering quality while requiring the least storage overhead.
\end{itemize}

\vspace{-4mm}
\section{Related work}

\paragraph{Head Avatar Animation.}
The animation generation of head avatars is divided into 3D and 2D scenes. In 3D animations, the FLAME model~\cite{li2017learning} is commonly used for generation and editing tasks. For instance, studies such as VOCA~\cite{VOCA2019}, COMA~\cite{ranjan2018generating}, FaceFormer~\cite{fan2022faceformer}, and SelfTalk~\cite{peng2023selftalk} utilize the FLAME model for speech-driven head avatar animation generation. Meanwhile, methods like EmoTalk~\cite{peng2023emotalk} and PiSaTalker~\cite{chen2023diffusiontalker} are based on blendshapes to construct 3D head avatars. In 2D animations, Gafni et al.~\cite{gafni2021dynamic} proposes a NeRF model based on expression vectors learned from monocular videos. HeadNeRF~\cite{hong2022headnerf} develop a parametric head model based on NeRF, using 2D neural rendering to enhance efficiency. INSTA~\cite{zielonka2023instant} deforms sampling points to canonical space by locating the nearest triangle on the FLAME mesh, achieving rapid rendering. 
Recent 3D Gaussian Splatting has been applied to dynamic head modeling.
GaussianAvatars~\cite{qian2024gaussianavatars} binds 3D Gaussians to a parametric morphable face model, allowing them to move with the dynamic mesh. 
GaussianHeadAvatar~\cite{xu2023gaussian} initializes 3D Gaussians with a neutral mesh head and then utilizes an MLP-based deformation net to capture complex expressions.
FlashAvatar~\cite{xiang2024flashavatar} attaches 3D Gaussians to the facial UV map, but the limited resolution of the 2D UV map restricts its ability to represent a dynamic 3D head and accurately capture complex facial topology.

However, the above methods require substantial and dynamic storage consumption, posing challenges for practical applications. Conversely, GraphAvatar stores only the GNN models instead of the 3D Gaussians, significantly reducing storage requirements. Additionally, we introduce a graph-guided optimization module and a 3D-aware enhancer to alleviate reliance on accurately tracked parameters.

\begin{figure*}[!ht]

    \centering
    \includegraphics[width=1.0\textwidth]{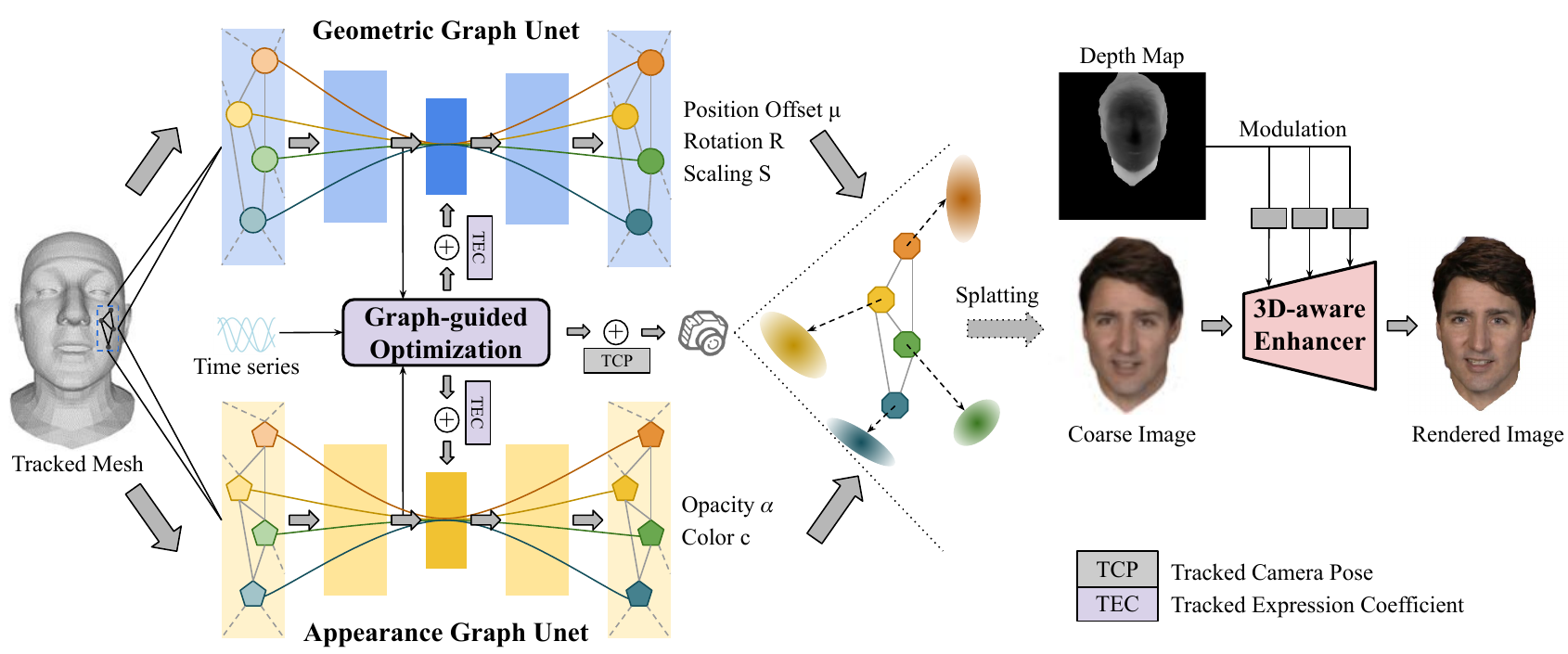}
    \vspace{-2mm}
    \caption{Pipeline of GraphAvatar. Our method takes the tracked meshes from source videos as input and first utilizes a geometric Graph Unet and an appearance Graph Unet to generate corresponding 3D Gaussian attributes. These Gaussians are then established as anchors to predict view-dependent attributes as neural Gaussians. To minimize errors from the tracked mesh, we introduce a graph-guided optimization module that utilizes time series and bottleneck features from Graph Unet to refine the tracked camera pose and expression coefficients. All Gaussians are combined and splatted into 2D images and depths using a differentiable rasterizer. Conditioned on the predicted depth map, a 3D-aware enhancer post-processes the rendered images to produce the final high-quality images.}
    \label{fig:pipeline}
    \vspace{-6mm}
\end{figure*}
\vspace{-2mm}
\section{Method}

\vspace{-2mm}
\subsection{Preliminaries}
Given a set of images of a static scene and the corresponding camera poses, 3DGS~\cite{kerbl20233d} learns a 3D scene representation using a set of 3D Gaussians to achieve novel view synthesis. 3DGS employs the point cloud obtained from structure from motion~\cite{schonberger2016structure} to initialize the position of 3D Gaussians. 
Each 3D Gaussian is defined as a tuple comprising covariance matrix $\Sigma \in \mathbb{R}^{3\times3}$, center $\mu \in \mathbb{R}^3$, view-dependent color $c \in \mathbb{R}^{3(k+1)^2}$, and opacity $\alpha \in \mathbb{R}$, denoted by $G = (\Sigma, \mu, c, \alpha)$, where $k$ is set to 3 representing the degree of the spherical harmonics.
The Gaussians are defined in the world coordinate, centered at the mean point as:

\vspace{-2.5mm}
\begin{equation}
    G(x) = e^{- \frac{1}{2}(x-\mu)^T\Sigma^{-1}(x-\mu)},
\end{equation}

To ensure stable training and guarantee that $\Sigma$ is positive semi-definite, the covariance matrix is further decomposed into rotation $R \in \mathbb{R}^4 $ and scaling $S \in \mathbb{R}^3 $:

\vspace{-2.5mm}
\begin{equation}
    \Sigma = RSS^{T}R^{T},
\end{equation}

The Gaussians are rasterized using a differentiable rasterizer, which 
projects them into image space. The pixel color $C$ is computed as:

\vspace{-5.5mm}
\begin{equation}
    C = \sum_{i=1}{c_i \alpha_i' \prod_{j=1}^{i-1}}(1-\alpha_j'),
\end{equation}
where $c_i$ is the color of each Gaussian. The blending weight $\alpha'$ is calculated by evaluating the 2D projection of the 3D Gaussian, which is then multiplied by the opacity $\alpha$. 
This process is efficiently executed by the differentiable rasterizer, resulting in successful image rendering.

\vspace{-2mm}
\subsection{GNN-based Avatar Representation}

Intuitively, we aim to learn a function $F(G(x)) = C$ that maps animatable 3D Gaussians into rasterized avatar images. However, rendering high-fidelity head avatars typically requires more than 10,000 Gaussian parameters. During the densification process, the number of Gaussians fluctuates with training, leading to dynamic storage requirements when modeling different head avatars with various facial expressions.
Therefore, our approach involves generating rather than directly optimizing the 3D Gaussians binding onto tracked meshes. 
To effectively capture the non-linear variations and diverse expressions in tracked meshes, employing a graph neural network is intuitive. This allows for the aggregation of complex geometric features inherent in facial topology.

Inspired by ~\cite{ranjan2018generating}, we utilize a geometric Graph Unet $U_{geo}$ and an appearance Graph Unet $U_{app}$ to capture non-linear topology for each tracked mesh $M = (V,A)$. A 3D facial mesh $M$ is a set of vertices $|V|=n, V \in \mathbb{R}^{n \times 3}$ and space adjacency matrix $A \in {0,1}^{n \times n}$, where $A_{ij} = 1$ denotes an edge connecting vertices $i$ and $j$. Since the adopted tracking method is FLAME-based, the number $n$ is 5023, corresponding to the vertex count of the FLAME template. The non-normalized graph Laplacian is defined as $L=D-A$, in which diagonal matrix $D$ represents the degree of each vertex. Since the Laplacian is diagonalized by the Fourier basis $U \in \mathbb{R}^{n \times n}$ as $L = U \Lambda U^T$, the convolution operator $*$ can be defined in Fourier space as a Hadamard product: $x * y = U((U^Tx)\odot(U^Ty))$. To reduce the computation, we select the Chebyshev graph convolution $g_{\theta_{i,j}}$ as the convolution layer in the Unet, which can be defined as:
\vspace{-3mm}
\begin{equation}
y_j = \sum_{i=1}^{F_{in}} g_{\theta_{i,j}}(L)x_{i},
\end{equation}
where $y_j$ indicates the output features of the Chebyshev graph convolution with trainable params $\theta_{i,j}$ and the input $x \in \mathbb{R}^{n \times F_{in}}$ has $F_{in}$ features for each vertex.
We further calculate the normals $N \in \mathbb{R}^{n \times 3}$ and concatenate them with the vertices to serve as input features. Thus, $F_{in}=6$ denotes the features of vertices, which include their position and normal values.
Except for the last layer, which has a different output dimension, $U_{geo}$ (geometric Unet) and $U_{app}$ (appearance Unet) consist of the same convolution layers. To avoid excessive smoothing caused by graph convolution, we utilize two graph convolution layers as the encoder, incorporating a mesh sampling operation as mentioned in ~\cite{ranjan2018generating}, to extract the bottleneck features $z \in \mathbb{R}^{8}$. Then we concat $z$ with expression coefficients $e$ and decode them using the same two graph convolution layers and upsample the graph back to the original number of vertices. Different activation functions are then applied to the vertex features, which are set as the attributes of the 3D Gaussians. The entire generation process is formulated as:
\vspace{-1mm}
\begin{align}
\delta_{\mu}, R, S &= \text{Act}(U_{geo}(X, A)), \\
c, \alpha &= \text{Act}(U_{app}(X, A))
\end{align}

where $X \in \mathbb{R}^{n \times 3}$ represents the position and normals for all vertices, and $A$ denotes the adjacency matrix for the tracked mesh $M$. The function Act refers to the activation function used in 3DGS, applied to the output vertex features and obtain Gaussians. The generated 3D Gaussians, oriented from the mesh vertices, possess attributes including $\{\mu', R, S, c, \alpha \}$. We introduce a learnable offset to the original vertex position to serve as the center of the generated Gaussians: $\mu' = \mu + \delta_{\mu}$.

Since the FLAME-based face model lacks vertices to describe hair and accessories, merely generating Gaussian parameters from the tracked mesh is insufficient to cover the entire space of a dynamic head. Inspired by Scaffold-GS~\cite{lu2023scaffold}, we treat the Gaussians generated by the Graph Unet as anchor points $G_{anc}$ and assign several neural Gaussians $\{G_0, ..., G_{k-1}\}$ binding to each anchor to represent the complex topology not included in the tracked mesh. For each anchor, we spawn $k$ view-dependent neural Gaussians and predict their attributes as follows:
\vspace{-2mm}
\begin{equation}
    \{\mu_0,...,\mu_{k-1}\} = \mathbf{\mu}+\{\delta_{\mu_0},\dots,\delta_{\mu_{k-1}}\} \cdot S,
\end{equation}
where $\{\delta_{\mu_0},\dots,\delta_{\mu_{k-1}}\} \in \mathbb{R}^3$ are the learnable offsets. For the other attributes of neural Gaussians, we decode them with individual MLPs. Take opacity $\alpha$ as an example, the process is formulated as:
\vspace{-2mm}
\begin{equation}
    \{\alpha_0, ..., \alpha_{k-1}\} = F_{\alpha}(f_{anc}, \vec{d}_{c}, e),
\end{equation}
where $\{\alpha_0, ..., \alpha_{k-1} \} \in \mathbb{R}^1 $ denotes the opacity for neural Gaussians. $F_{\alpha}$ denotes the MLPs used to generate the corresponding attribute. For the input, $f_{anc}$ is the learnable anchor features, $\vec{d}_{c}$ is the direction between the camera and the anchor Gaussian and $e$ is expression coefficients from face tracking. Other attributes can be obtained similarly. 

Finally, we gather all the attributes from anchor Gaussians and neural Gaussians as $G=\{G_{anc}, G_0, ..., G_{k-1}\}$, which are rasterized into coarse 2D images $I_{c}$ and depth maps $D$.
\vspace{-2mm}
\subsection{Graph-guided Optimization}

To mitigate the inaccuracies stemming from tracked FLAME parameters, including camera poses and expression coefficients, we have developed a graph-guided optimization module (GGO) to refine these parameters throughout the training process. Inspired by ~\cite{ming2024high} that introduces a temporal regressor to rectify coefficients and ensure smoothness, we input the normalized time $t$ and process it through MLPs to extract temporal features $f_t$. Subsequently, we concatenate the bottleneck features from both Graph Unets, denoted as $f_g$. Upon generating these features, we execute a cross-attention mechanism (Attn) between $f_t$ and $f_g$, which enables the prediction of offsets for the tracked parameters. It can be formulated as:
\vspace{-2mm}
\begin{equation}
    \delta=\text{Attn}(q=f_g,k=f_t,v=f_t), f_t=\text{MLP}(t)
\end{equation}
Then we divide the prediction into two components: $\delta_{e}$ and $\delta_{pose}$, which represent the offsets for the tracked expression coefficients and the camera pose offsets within the Lie group $SO(3)$, denoting the space of 3D rotations. Subsequently, we transform $\delta_{pose}$ into a transformation matrix and apply these offsets to the original tracked parameters. Through such a process, GraphAvatar can optimize both facial expressions and camera poses guided by the graph features in an end-to-end manner.
\vspace{-2mm}
\subsection{3D-aware Enhancer}
To achieve higher rendering quality, we have designed a 3D-aware enhancer specifically for detail restoration. Instead of merely concatenating the rendered maps as the input to a Unet post-processor, we treat the depth signal $D$ separately. This depth signal is integrated into every block of the Unet through a learned transformation that modulates the activation within each block. This method allows for a more nuanced adjustment based on depth information, enhancing the details of rendered images. 

Formally, let $F^k$ represent the activation of an intermediate block within the network, where $k$ indicates the block level and $C_k$ is the channel dimension at that level. The depth map $D$ undergoes a transformation, such as a $1 \times 1$ convolution, to predict scale $\gamma_{i,j}^k$ and bias $\beta_{i,j} $ values, which match the channel dimension $C_k$. These scale and bias values are then used to modulate the activations $F^k$ according to the following formula:
\vspace{-1.5mm}
\begin{equation}
\tilde{F}^k_{i,j} = \gamma_{i,j}^k(D) \odot F^k_{i,j} + \beta_{i,j}(D).
\end{equation}
where $\odot$ denotes element-wise product, $i$ and $j$ indicate the spatial position. Integrating the local information from nearby pixels has proven effective for recovering high-frequency details, particularly in dynamic avatar animations. Learning local correlations facilitates the extraction of patterns across spatial regions, linking them closely to the underlying 3D geometric structure. Additionally, modulating the process with depth maps introduces geometric guidance that further regularizes the learning process. The proposed 3D-aware enhancer ensures that the enhancements are not only visually compelling but also geometrically coherent, leading to high-fidelity facial animation.

\begin{table*}[!ht]
    \centering
    \caption{Results on the dataset provided by INSTA~\cite{Zielonka2022InstantVH} and NeRFBlendShape (NBS)~\cite{Gao2022nerfblendshape}. We report L2 distance, PSNR, SSIM, LPIPS, inference time (rendering time for one frame) and model size. We highlight the best performance in bold. Ours GraphAvatar achieves the highest quality in avatar animation rendering with competitive inference time and minimal model storage size.}

    \resizebox{1.0\textwidth}{!}{%
    \begin{tabular}{l|c|cccc|cc} 
    \toprule
         Method & dataset & L2 $\downarrow$ & PSNR $\uparrow$ & SSIM $\uparrow$ &  LPIPS $\downarrow$ & Time (s) $\downarrow$ & Size (MB) $\downarrow$ \\
         \midrule
         NHA~\cite{grassal2022neural} & \multirow{7}{*}{INSTA} & 0.0024 & 26.99 & 0.942 & 0.043 & 0.63 & 90.8 \\
         IMAvatar~\cite{zheng2022avatar} & & 0.0021 & 27.92 & 0.943 & 0.061 & 12.34 & 22.7 \\
         INSTA~\cite{Zielonka2022InstantVH} &  & 0.0017 & 28.61 & 0.944 & 0.047 & 0.052 & 523.7 \\
         FlashAvatar~\cite{xiang2024flashavatar} &  & 0.0017 & 29.20 & 0.949 & 0.040 & 0.007 & 13.3 \\
         Gaussian Head Avatar~\cite{xu2023gaussianheadavatar} &  & 0.0016 & 29.71 & 0.958 & 0.043 & 0.011 & 43.1 \\
         GaussianAvatars~\cite{qian2024gaussianavatars} &  & 0.0014 & 29.06 & 0.953 & 0.046 & \textbf{0.006} & 71.3 \\
         \methodNameSpace (Ours) &  & \textbf{0.0011} & \textbf{30.52} & \textbf{0.966} & \textbf{0.036} & 0.029 & \textbf{10.8} \\
         \midrule
         NeRFBlendShape~\cite{Gao2022nerfblendshape} & \multirow{5}{*}{NBS} & 0.0035 & 24.81  & 0.935  & 0.086 & 0.10 & 538.7 \\ 
         FlashAvatar~\cite{xiang2024flashavatar} &  & 0.0028 & 25.68 & 0.939 & 0.073 & 0.007 & 13.3 \\
         Gaussian Head Avatar~\cite{xu2023gaussianheadavatar} &  & 0.0022 & 26.38 & 0.944 & 0.064 & 0.011 & 43.8 \\
         GaussianAvatars~\cite{qian2024gaussianavatars} &  & 0.0024 & 25.56 & 0.937 & 0.073 & \textbf{0.006} & 42.7 \\
         \methodNameSpace (Ours) &  & \textbf{0.0018} & \textbf{27.32} &  \textbf{0.953} & \textbf{0.061} & 0.029 & \textbf{10.8} \\
         \midrule

    \end{tabular}
    }
    \vspace{-5mm}
    \label{tab:main_result}
\end{table*}

\vspace{-2mm}
\subsection{Training}
Given the complexities involved in optimizing graph neural networks, we initiate with a warm-up stage for parameter initialization. We treat the target actor as a static scene and utilize vanilla 3D Gaussian Splatting (3DGS) to produce pseudo Gaussians. Subsequently, we implement an MSE (Mean Squared Error) loss between the Gaussians $G_{anc}$ generated by the Graph Unet and the pseudo Gaussians. This warm-up phase is rapid, consisting of 10,000 iterations.

After the initialization of the graph neural networks, we proceed to train the full pipeline of GraphAvatar. We supervise the rendered image $I_f$ using a combination of L1 loss, SSIM (Structural Similarity Index) loss, and LPIPS (Learned Perceptual Image Patch Similarity) loss. These loss functions help in refining the final image quality by focusing on both pixel-level accuracy and perceptual similarity, thereby ensuring that the rendered images are both visually accurate and aesthetically pleasing:
\vspace{-1.0mm}
\begin{equation}
\mathcal{L}_{f}=(1-\lambda)\mathcal{L}_1+\lambda \mathcal{L}_{D-SSIM}+\lambda_{LPIPS} \mathcal{L}_{LPIPS}
\end{equation}
with $\lambda=0.2, \lambda_{LPIPS}=0.1$. To maintain the render quality of coarse images $I_c$, we also supervise it with L1 loss and SSIM loss, which is denoted as $\mathcal{L}_c$. Additionally, to restrict the growing space for neural Gaussians, GraphAvatar further renders weight maps to denote the probability between foreground and background. We supervise the weight maps against the foreground head segmentation using cross-entropy loss, denoted as $\mathcal{L}_w$. Our final loss function is:
\vspace{-1.0mm}
\begin{equation}
\mathcal{L}=\lambda_f \mathcal{L}_{f}+\lambda_c \mathcal{L}_{c} + \lambda_w \mathcal{L}_{w}
\end{equation}
where $\lambda_f=1.0$, $\lambda_c=0.1$ and $\lambda_w=0.1$. GraphAvatar is trained with the final loss until convergence.

\begin{figure*}[!ht]
    \centering
    \includegraphics[width=0.85\textwidth]{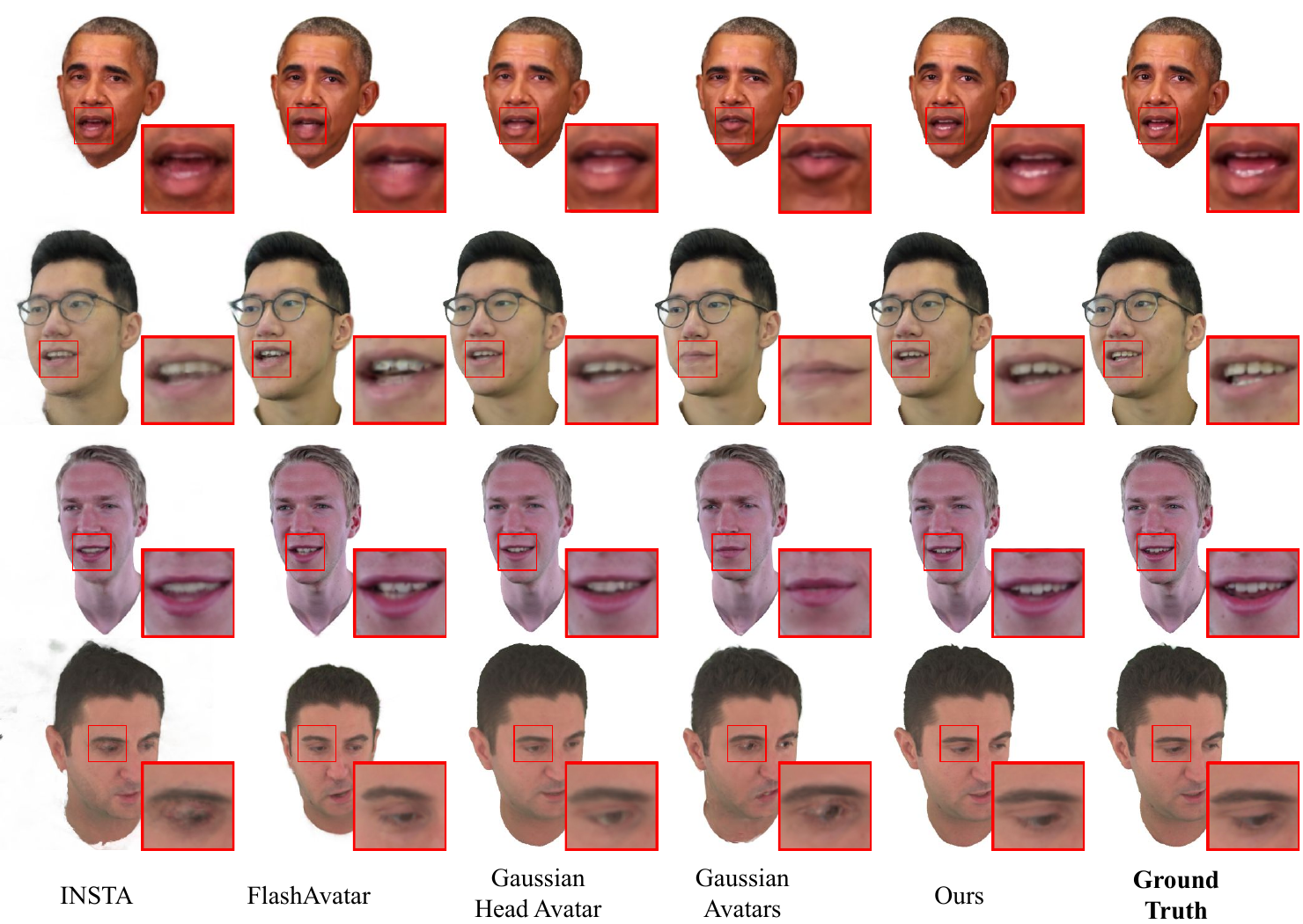}
    \vspace{-2mm}
    \caption{Qualitative comparison on INSTA dataset. }
    \label{fig:insta_qua}
    \vspace{-4mm}
\end{figure*}

\begin{figure*}[!t]
    \centering
    \includegraphics[width=0.85\textwidth]{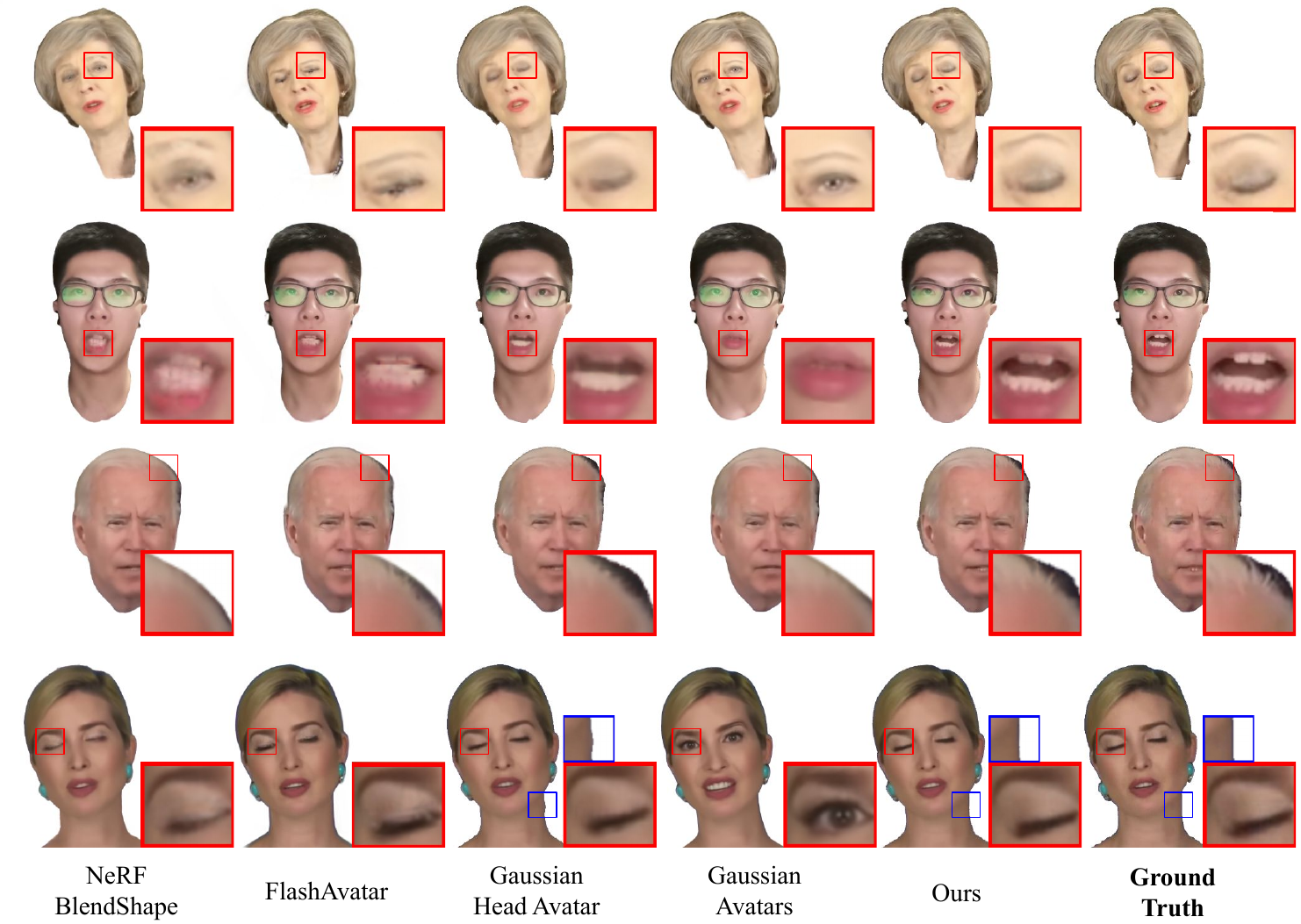}
    \vspace{-2mm}
    \caption{Qualitative comparison on NBS dataset. }
    \label{fig:nbs_qua}
    \vspace{-4mm}
\end{figure*}
\vspace{-2mm}

\section{Experiments}

\subsection{Experimental Settings}
\label{sec:4.1}

\paragraph{Datasets and Baselines. } 
We evaluate our model using two challenging datasets: NeRFBlendShape (abbreviated as NBS data) and INSTA. The NBS dataset~\cite{Gao2022nerfblendshape} comprises monocular videos from eight subjects, with the last 500 frames of each subject's video designated as the test set. Similarly, the INSTA dataset~\cite{Zielonka2022InstantVH} includes data from ten subjects, with the final 350 frames of each sequence reserved for testing.

For the INSTA dataset, we select NeRF-based methods including NHA~\cite{grassal2022neural}, IMAvatar~\cite{zheng2022avatar}, and INSTA~\cite{Zielonka2022InstantVH}, as well as 3DGS-based methods such as FlashAvatar~\cite{xiang2024flashavatar}, Gaussian Head Avatar~\cite{xu2023gaussianheadavatar}, and GaussianAvatars~\cite{qian2024gaussianavatars}. 
Since FlashAvatar and GaussianAvatars outperform the previous point-based method PointAvatar~\cite{zheng2023pointavatar}, we opted to compare with more advanced methods instead.
To ensure a fair comparison, we adhere to the data preprocessing protocols provided by each baseline. However, since several tracking data required by the 3DGS-based methods are absent in the originally provided dataset, we retrack the subjects using the metrical-tracker~\cite{zielonka2022towards}.

For the NBS dataset, we select NeRF-based methods including NeRFBlendShape~\cite{Gao2022nerfblendshape} and 3DGS-based methods such as FlashAvatar~\cite{xiang2024flashavatar}, Gaussian Head Avatar~\cite{xu2023gaussianheadavatar}, and GaussianAvatars~\cite{qian2024gaussianavatars}. To ensure a fair comparison, we mask the background in white and maintain the head and neck of the subjects as the foreground. Since NeRFBlendShape is initially trained only for the head, we retrain it to render avatars that include both the head and neck until full convergence. Additionally, the NBS dataset, based on NeRFBlendShape, does not provide FLAME-based tracking parameters. Thus, we retrack all subjects using the metrical-tracker~\cite{zielonka2022towards} and store tracking results for the training of 3DGS-based methods.
\vspace{-2mm}
\paragraph{Metrics. }
To assess the quality of the synthesized images, we report common metrics such as Mean Squared Error (L2), Peak Signal-to-Noise Ratio (PSNR), Structural Similarity (SSIM) and Learned Perceptual Image Patch Similarity (LPIPS)~\cite{zhang2018unreasonable}. Additionally, we report the inference time for rendering in seconds and the model size for storage occupation in megabytes. 


\vspace{-2mm}
\subsection{Head Avatar Animation}
\label{sec:4.2}

\paragraph{Quantitative comparison.} In this experiment, we animate the learned actors using novel poses and expression parameters derived from their test sets. 

Tab.~\ref{tab:main_result} displays the metric comparisons against baseline methods, with separate sections for the INSTA and NBS datasets. It is evident from the table that NeRF-based methods, such as NeRFBlendShape and INSTA, yield significantly larger model sizes compared to the 3DGS-based method, underscoring the potential of the 3DGS method for model compression.
More notably, GraphAvatar surpasses recent 3DGS-based baselines in terms of both rendering quality and model size. This validates our approach of replacing 3D Gaussians with GNNs to store parameters, which not only compacts the model size but also preserves performance. 
FlashAvatar assigns Gaussians based on the 2D facial UV map, but the limited resolution leads to ambiguous and unclear details in the UV map.
Compared to Gaussian Head Avatar and GraphAvatar, which rely on face tracking, GraphAvatar uses a graph-guided optimization strategy that effectively reduces the accumulation of prior knowledge errors and decreases dependency on tracked meshes.
The 3D-aware enhancer in GraphAvatar significantly enhances image quality, contributing to its top PSNR scores. 

\paragraph{Qualitative comparison.} Fig.~\ref{fig:insta_qua} and Fig.~\ref{fig:nbs_qua} provide qualitative comparisons with the latest baseline methods for each dataset. 
Considering the FLAME model's inability to model teeth accurately, we primarily compared the image rendering precision of various methods on the dental region as shown in the first three rows of Fig.~\ref{fig:insta_qua}. Other methods, overly reliant on tracked meshes or UV maps, failed to capture the dental details accurately. In contrast, our method utilizes a graph-guided optimization strategy that significantly reduces the accumulation of prior knowledge errors during training, thus minimizing dependency on tracked meshes. This allows for better rendering of intricate tooth structures. In the fourth row, we selected an extreme angle as a test case, where other methods showed artifacts to varying degrees, failing to render facial features like eyes accurately. Notably, FlashAvatar, to reduce the number of 3D Gaussians, relies on UV maps, losing vital 3D preparatory knowledge, resulting in the largest angle errors in generating head avatars. Conversely, our method accurately produces head avatars even at extreme angles, closely matching the ground truth.

In Fig.~\ref{fig:nbs_qua}, our method achieves optimal image rendering precision in details like eyes, mouth, teeth, and hair strands. GaussianAvatars struggles to accurately control the closure of eyes and mouths, possibly due to its poor sensitivity to expression coefficients and greater dependence on tracked meshes. In the fourth row, our method's image rendering precision for the eye region is close to that of Gaussian Head Avatar, but the latter occasionally shows flaws, as indicated by the blue box. In summary, it can be seen that GraphAvatar preserves the best details. Please refer to the supplementary material for more visualization results.

\vspace{-3mm}
\subsection{Component-wise Ablations}

In this section, we conduct a component-wise ablation study to elucidate the influence of each component. Initially, we train GraphAvatar from scratch without the warm-up stage, denoted as ``w/o Warm-up Stage", to evaluate the influence of pseudo Gaussians for graph optimization. 
Next, we omit the neural Gaussian, referring to this as ``w/o Neural Gaussian", which retains only the anchor points generated by the Graph Unet. This allows us to evaluate the impact of the neural Gaussian on the FLAME-based face model's missing elements, such as hair and accessories.
Finally, we exclude the Graph-guided Optimization module and the 3D-aware Enhancer in the method section to assess the contributions of these proposed components.

As shown in Tab.~\ref{tab:ablation}, our GraphAvatar with the full pipeline achieves the best performance. Without the warm-up stage, GraphAvatar exhibits unstable training and difficulty converging, which is attributed to the characteristics of the graph neural network. 
Without the neural Gaussian, GraphAvatar is constrained by the FLAME template, lacking the ability to model hair and accessories, making it difficult for anchor Gaussians to cover the entire 3D head. Without the Graph-guided Optimization module, accumulation errors from FLAME coefficients and tracking camera poses lead to a decline in rendering quality. The absence of the 3D-aware Enhancer results in over-smoothness introduced by the GNN, which negatively impacts human-perceptual metrics like SSIM and LPIPS. The two graph Unets consume the majority of the time and model size, while the other three components occupy only a small portion but play crucial roles in different aspects. GraphAvatar achieves an optimal trade-off between rendering quality and model size.

\begin{figure}[!t]
    \centering
    \includegraphics[width=1.0\linewidth]{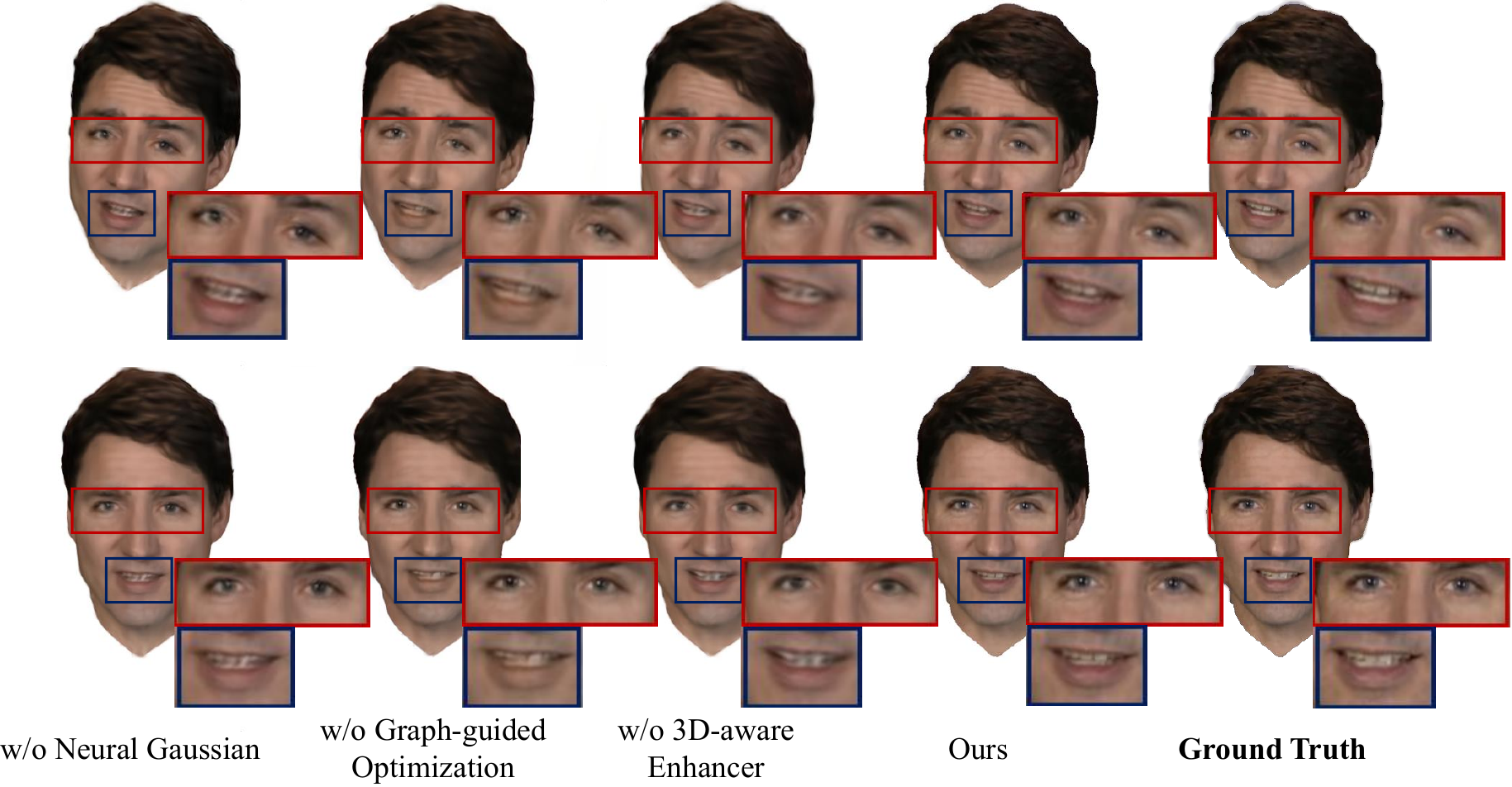}
    \vspace{-4mm}
    \caption{Qualitative ablation study on the INSTA dataset. 
    }
    \label{fig:ablation}
    \vspace{-4mm}
\end{figure}

To further illustrate the influence of each technique, we also implement a visual ablation in Fig.~\ref{fig:ablation}. Without the extension of neural Gaussian and graph-guided optimization, GraphAvatar is constrained by the tracked mesh, leading to blurred hair that is not included in the tracked mesh transformed from the FLAME template. Additionally, due to the over-smooth characteristics of GNN, the network tends to generate coarse images with fewer details. Thus, we introduce a 3D-aware enhancer to capture high-frequency details and refine the rendered images. It can be observed that with all these techniques, the rendered images retain more details in the eyes and hair, leading to improvements in SSIM and LPIPS scores that better align with human visual perception metrics. GraphAvatar strikes a better balance between achieving optimal visual quality and maintaining a lightweight model. Please refer to the supplementary materials for more ablation studies in different settings.

\begin{table}[!t]
    \centering
    \caption{Component-wise ablation study of \methodNameSpace on the INSTA dataset.}
    \vspace{-2mm}
    \resizebox{1.0\linewidth}{!}{%
    \begin{tabular}{l|c@{\hskip 0.1in}c@{\hskip 0.1in}c@{\hskip 0.1in}c@{\hskip 0.1in}|c@{\hskip 0.1in}c@{\hskip 0.1in}c}
    \toprule
    Method   & L2 $\downarrow$  & PSNR $\uparrow$ & SSIM $\uparrow$ & LPIPS $\downarrow$ & Time (s) $\downarrow$ & Size (MB) $\downarrow$ \\
    \midrule
    w/o Warm-up Stage              & 0.1283 & 8.97  & 0.818  & 0.334 & 0.029 & 10.8 \\
    w/o Neural Gaussian            & 0.0014 & 29.61 & 0.945  & 0.051 & 0.027 & 10.5 \\
    w/o Graph-guided Optimization  & 0.0012 & 30.41 & 0.963  & 0.041 & 0.028 & 10.6 \\
    w/o 3D-aware Enhancer          & 0.0011 & 30.50 & 0.959  & 0.042 & 0.024 & 10.1 \\
    Full Pipeline                          & \textbf{0.0011}  & \textbf{30.52} & \textbf{0.966} & \textbf{0.036} & 0.029 & 10.8 \\
     \bottomrule
    \end{tabular}
    }
    \label{tab:ablation}
    \vspace{-5mm}
\end{table}

\vspace{-3mm}
\section{Conclusion}

In this work, we propose GraphAvatar, a compact method that takes tracked meshes as input and uses graph neural networks to generate the parameters of 3D Gaussians, ultimately rendering dynamic avatar animations. 
GraphAvatar employs Graph Unets, significantly reducing the storage consumption compared to directly storing 3D Gaussians. 
Our method achieves state-of-the-art performance in both image quality and storage consumption, opening up new possibilities for advanced digital human avatar applications.
\vspace{-4mm}
\section{Acknowledgments}
This work was supported by the National Key R\&D Program of China (2022ZD0117900), the NSFC (62332015), the Beijing Natural Science Foundation (L247008), the project of China Disabled Persons Federation (CDPF2023KF00002), and the Key Project of ISCAS (ISCAS-ZD-202401).

\bibliography{aaai25}

\end{document}